\documentclass[runningheads]{llncs}
\usepackage[T1]{fontenc}
\usepackage{graphicx}
\usepackage{hyperref}
\usepackage{color}

\usepackage{times}
\usepackage{epsfig}
\usepackage{amsmath}
\usepackage{amssymb}
\usepackage{bm}
\usepackage{booktabs}
\usepackage{tablefootnote}
\usepackage{color}

\newcommand{\JZ}[1]{#1}

\newcommand{\paragraphdata}[1]{{\textbf{#1}}}

\begin{document}
\title{Precise Location Matching Improves Dense Contrastive Learning in Digital Pathology}
% Precise Location-based Matching Improves Self-Supervision in Digital Pathology}

%%%%%%%%%%%%%% use for submission
% \author{Anonymous}
% \authorrunning{Anonymous et al.}
% \institute{Anonymous Organization\\
% { \email{**@******.***} } }

\newcommand*\samethanks[1][\value{footnote}]{\footnotemark[#1]}

%%%%%%%%%%%%% author list
\author{Jingwei Zhang\inst{1}\thanks{These authors contributed equally to this paper.} \and
Saarthak Kapse\inst{1}\samethanks[1] \and Ke Ma \inst{2} \and Prateek Prasanna\inst{1} \and \\
Maria Vakalopoulou\inst{3} \and
Joel Saltz\inst{1} \and
Dimitris Samaras\inst{1}}
\authorrunning{J. Zhang et al.}
\institute{Stony Brook University, USA \and Snap Inc., USA
\and
CentraleSupélec, University of Paris-Saclay, France\\
{ \email{\{jingwezhang, kemma, samaras\}@cs.stonybrook.edu}} \\
 \email{\{saarthak.kapse, prateek.prasanna\}@stonybrook.edu} \\
\email{maria.vakalopoulou@centralesupelec.fr
Joel.Saltz@stonybrookmedicine.edu} }

\maketitle              % typeset the header of the contribution
\begin{abstract} 
Dense prediction tasks such as segmentation and detection of pathological entities hold crucial clinical value in computational pathology workflows. However, obtaining dense annotations on large cohorts is usually tedious and expensive. 
Contrastive learning (CL) is thus often employed to leverage large volumes of unlabeled data to pre-train the backbone network. 
To boost CL for dense prediction, some studies have proposed variations of dense matching objectives in pre-training.
However, our analysis shows that employing existing dense matching strategies on histopathology images enforces invariance among incorrect pairs of dense features and, thus, is \textit{imprecise}.
To address this, we propose a \textit{precise location-based matching mechanism} that utilizes the overlapping information between geometric transformations to precisely match regions in two augmentations.
Extensive experiments on two pretraining datasets (TCGA-BRCA, NCT-CRC-HE) and three downstream datasets (GlaS, CRAG, BCSS) highlight the superiority of our method in semantic and instance segmentation tasks.
Our method outperforms previous dense matching methods by up to 7.2\% in average precision for detection and 5.6\% in average precision for instance segmentation tasks.
Additionally, by using our matching mechanism in the three popular contrastive learning frameworks, MoCo-v2, VICRegL, and ConCL, the average precision in detection is improved by 0.7\% to 5.2\%, and the average precision in segmentation is improved by 0.7\% to 4.0\%, demonstrating generalizability. Our code is available at \href{https://github.com/cvlab-stonybrook/PLM_SSL}{https://github.com/cvlab-stonybrook/PLM\_SSL}

\keywords{Dense contrastive learning \and Self-supervised learning \and Segmentation \and Detection \and Computational Pathology}
\end{abstract}

\section{Introduction}

% Computational pathology is becoming increasing popular in modern healthcare. 
% More pathology images, including Whole Slide Images, are now analysed by deep learning (DL) techniques~\cite{zhang2022gigapixel,zhang2021joint,wang2019pathology,tellez2019neural}.
In computational pathology, dense prediction tasks such as segmentation and detection are essential in analyzing digitized histology scans~\cite{tellez2019neural,wang2019pathology}.
However, unlike classification, obtaining labels from pathologists for dense prediction tasks is very tedious and expensive. 
%Currently, there are some publicly available datasets for dense prediction tasks such as \cite{gamper2020pannuke,graham2019hover} however, the generalization of models trained on them is not guaranteed due to the intra and inter-hospital variability~\cite{tellez2019quantifying}. 
%This leads to fewer ground truth annotations
% Especially in the pathology domain where expert is required for labeling.
% It usually result in a much fewer annotations in dense prediction task in this area.

% is a type of self-supervised learning (SSL) method that  learns representations of images  using large volumes of unlabeled data.
Contrastive learning (CL) is being increasingly adopted as a self-supervised learning (SSL) strategy in computational pathology~\cite{dsmil,hipt,boyd2021self} to reduce the need for annotations.
% It is a common method to alleviate the problem of few labels \cite{xxx}.
In standard CL, two augmented views are obtained from the input image, and the key idea is to pull the representations of these views closer while pushing apart representations from any other image. Popular CL methods such as SimCLR~\cite{chen2020simple_simclr}, MoCo~\cite{he2020momentum_moco_v1,chen2020improved_mocov2,chen2021empirical_movov3}, BYOL~\cite{grill2020bootstrap_byol}, and VICReg~\cite{bardes2022vicreg},
generalize well to multiple computer vision and medical imaging tasks.
% Because of the contrastive way of training, where the augmentation is conducted on the entire image, most existing CL methods have focused on global representations \cite{chen2020simple_simclr,chen2020improved_mocov2,grill2020bootstrap_byol}. 
% For each image, only one feature vector is generated to represent its entity and the training is performed using contrastive losses on this vector. Such an approach works pretty well on classification tasks that they usually rely on the the global representations of a given image. 
% \textbf{Drawbacks in current CL strategies.} 
In CL, traditionally, a spatial pooling operation is applied to the output feature map of the backbone network to encode each view into a global representation. These global CL approaches~\cite{chen2020simple_simclr,chen2020improved_mocov2,grill2020bootstrap_byol} work well when the downstream tasks involve classification problems; however, for dense prediction tasks, such representations are not optimal since they require detailed local descriptors of the images. Towards this direction, DenseCL~\cite{wang2021dense_densecl} and VICRegL~\cite{bardes2022vicregl} propose incorporating local details in pre-training through leveraging dense matching objectives between the feature maps across both views. In particular, representations from local patches are extracted, and their correspondences across the different views are investigated through the dense matching operation. 
%The representations in feature maps corresponds to small local patches in the input views. 
%For this dense matching, correspondence between these local patches or their corresponding representations across the view is required. 
DenseCL utilizes feature space cosine similarity matching (denoted by $M_{ft}$) between local representations across the views to find the correspondence pairs. Whereas VICRegL employs spatial location of local patches of the feature maps (denoted by $M_{loc}$) to find the closest spatial distance between corresponding pairs across the views. 
% The spatial location for each view is stored with reference to geometric augmentations on the input image. 

\begin{figure}[ht]
\begin{center}
\includegraphics[width=0.7\linewidth]{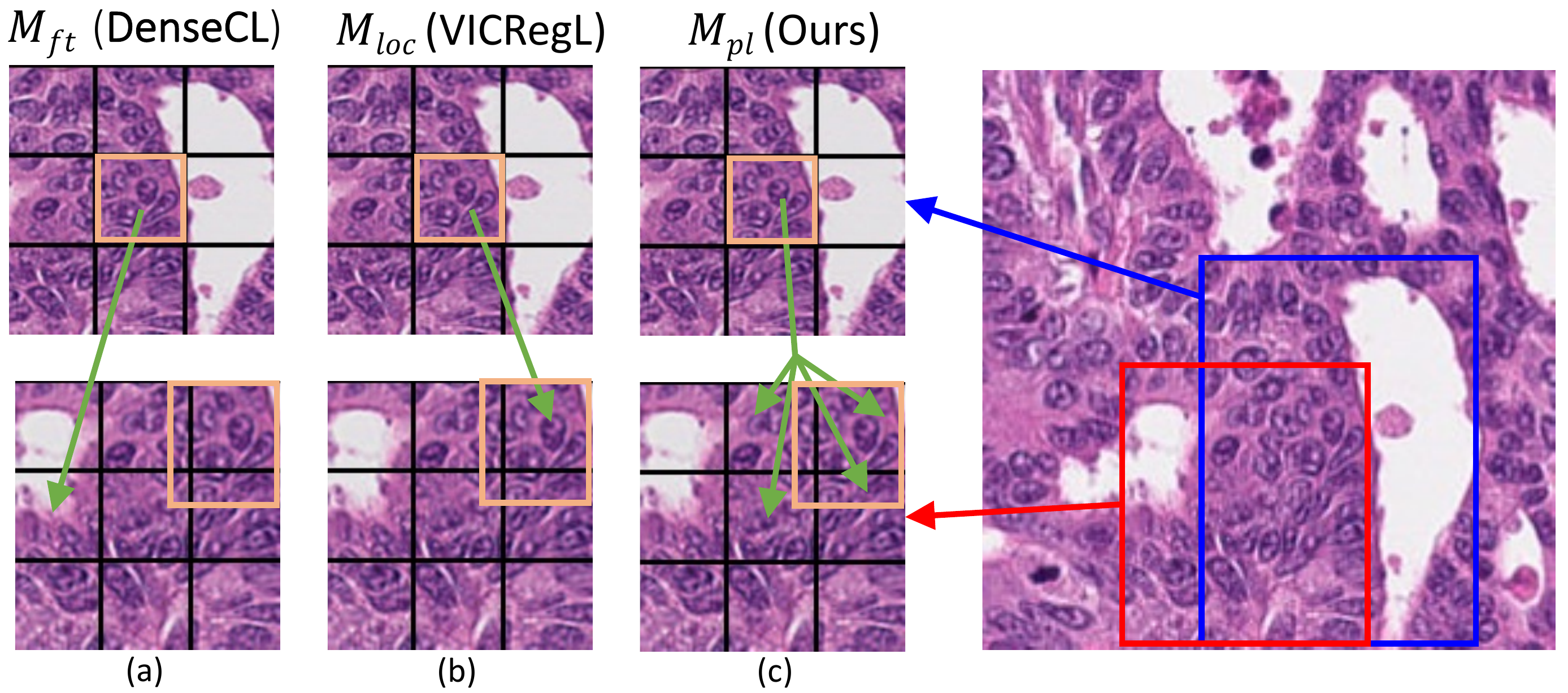}
\end{center}
   \caption{Local feature matching using three methods for two random augmentations of the same image. Green arrows indicate the matching operation, and orange boxes indicate the same regions in the two views.
       (a) \textbf{Feature similarity-based matching} ($M_{ft}$), used by DenseCL~\cite{wang2021dense_densecl}. The network erroneously pairs patches with multiple nuclei to a local patch containing mainly stroma and non-tissue.
       (b) \textbf{Location-based matching} ($M_{loc}$), proposed by VICRegL~\cite{bardes2022vicregl}. It matches a patch consisting of multiple nuclei to only one patch in another view containing fewer nuclei due to zooming augmentation. 
       (c) \textbf{Precise location-based matching} ($M_{pl}$). We match a patch to multiple patches in the other view by incorporating exact overlapping weights between the orange boxes across the views.
   }
\label{fig:matching_teaser}
\end{figure}

\textbf{Need for precise matching in pathology:} The pitfalls of both $M_{ft}$ and $M_{loc}$ can be observed in Fig. \ref{fig:matching_teaser} (a) \& (b) respectively. The feature-based matching, $M_{ft}$, erroneously matches the local patch consisting of multiple nuclei to a patch in another view predominantly consisting of stroma and non-tissue regions.
\JZ{This is because their similarity is defined on the features, which is significantly affected by the model.}
This could critically hamper the representation learning in histopathology as the invariance between these two patches could force the model to focus on stroma-based descriptors while ignoring other crucial information about cells and their morphology. 
Location-based matching $M_{loc}$ avoids this error by storing the location information after the geometric transformations to find the pairs. However it can be observed in Fig. \ref{fig:matching_teaser} (b) that the local patch consisting of multiple nuclei is matched with a patch in the other view containing fewer nuclei.
\JZ{This is because they allow matching to strictly one patch.}
This invariance could potentially enforce the network to ignore crucial information regarding cell density. Due to the \textit{zooming} and \textit{cropping} augmentations, a local patch in a given view may overlap with multiple local patches in another. Formulation of $M_{loc}$ thus has the undesired constraint that a local patch in a view can only match to one corresponding local patch in the other, which is not precise and sub-optimal. Since histopathology images consist of numerous fine-grained individual entities/objects, there is a need for more precise dense matching across the views to overcome the limitations encountered and provide better representations for dense prediction tasks.

To this end, we propose a precise location-based matching strategy, denoted by $M_{pl}$, which matches a local patch in a view to multiple corresponding overlapping patches in another, as shown in Fig. \ref{fig:matching_teaser} (c). By relaxing the previous constraint, $M_{pl}$ enables \textit{precise matching} between the views. We demonstrate the efficacy of our precise matching strategy in dense prediction tasks involving detection and segmentation on multiple datasets across colon and breast cancer.
% To evaluate our method, we pre-train a ResNet backbone on the NCT-CRC-HE-100K~\cite{nct} and TCGA-BRCA~\cite{brca} datasets with the proposed precise location-based matching CL scheme, and apply the pre-trained backbone to two gland instance segmentation downstream tasks, GlaS~\cite{sirinukunwattana2017gland} and CRAG~\cite{graham2019mild}, and a breast semantic segmentation downstream task BCSS~\cite{bcss}.
Experiment results show that our precise location-based matching outperforms previous local matching strategies, improving average precision by up to 7.2\% for detection and 5.6\% for instance segmentation.
We further demonstrate the generalizability of our approach by adopting $M_{pl}$ in three popular contrastive learning frameworks: MoCo-v2~\cite{chen2020improved_mocov2}, VICRegL~\cite{bardes2022vicregl}, and ConCL~\cite{yang2022concl}.
$M_{pl}$ shows a relative improvement in average precision by 5.2\%, 1.5\%, 0.7\% for detection, and by 4.0\%, 2.9\%, 0.7\% for instance segmentation with the three aforementioned CL frameworks, respectively.

% 5.2 % and 9 % in detection, 4.0 % and 7.3 % in instance segmentation on GlaS and
% CRAG respectively. For the semantic segmentation task, compared to vanilla MoCo-v2,
% relative improvement in Jaccard score is 0.45 

% more constrained, can only match one corresponding pair across the view. Our method relaxes that constraint. Previsous constrain leads to incorrect local correspondances. We go step beyond and with our relaxed constraint our matching is precise. 

% applies CL matching between one feature representation and the weighted sum of all other representations that have a spatial overlap.
%maps one feature in a augmentation to the weighted sum of all features in another augmentation which have an overlapping region with this feature.

% To evaluate our method, we pre-train a ResNet backbone on the NCT-CRC-HE-100K~\cite{nct} and TCGA-BRCA~\cite{brca} datasets with the proposed precise location-based matching CL scheme, and apply the pre-trained backbone to two gland instance segmentation downstream tasks, GlaS~\cite{sirinukunwattana2017gland} and CRAG~\cite{graham2019mild}, and a breast semantic segmentation downstream task BCSS~\cite{bcss}.

% more constrained, can only match one corresponding pair across the view. Our method relaxes that constraint. Previsous constrain leads to incorrect local correspondances. We go step beyond and with our relaxed constraint our matching is precise. 

\section{Method}

\begin{figure}[ht]
\begin{center}
%\fbox{\rule{0pt}{2in} \rule{.9\linewidth}{0pt}}
\includegraphics[width=\linewidth]{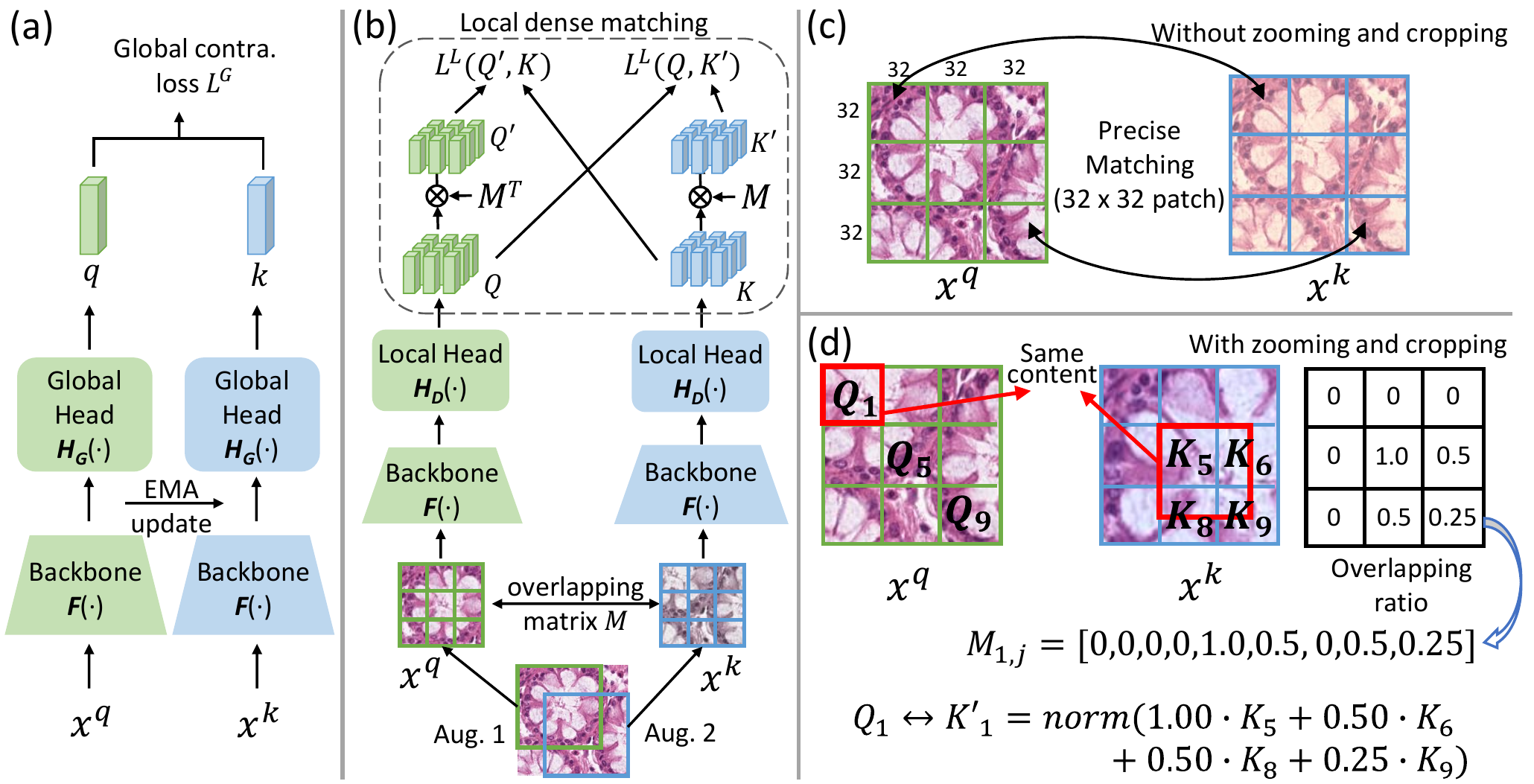}
\end{center}
   \caption{Overview of the proposed method.
   (a) Overall structure of our global contrastive framework, the same as MoCo-v2~\cite{chen2020improved_mocov2}. For each view of the image, we generate a single global feature representation that represents the entire image.
   (b) Overall structure of our local dense contrastive framework. For each view of the image, we generate several local feature representations. Each local feature represents a local patch in the image.
   (c) Without zooming and cropping, patches from two augmentations (for e.g., color jitter) precisely match with each other.
   (d) If the augmentation contains zooming and cropping, $Q_1$ matches the weighted sum of $K_5$, $K_6$, $K_8$ and $K_9$. The weights are calculated as the overlapping ratio of $K_i$ and the red boxed region in $x^k$ corresponding to $Q_1$. $M_{1, j}$ is the first row of overlapping matrix $M$. 
   %Note the figures are for illustration, the actual number of features ($|Q|$ and $|K|$) is much larger.
   }
\label{fig:framework}
\end{figure}

% \PP{Include a Notations section and a small paragraph on what you'll be discussing in this section. What are the main modules? How do they fit/interact with each other - a very high level description is first needed}

% Our method is based on MoCo-v2~\cite{he2020momentum_moco_v1,chen2020improved_mocov2}.
% However other CL frameworks could also be applied.
Our method consists of a global contrastive learning part similar to MoCo, which learns a global feature representation of an input image, and a local dense contrastive learning part that learns the local feature representations of small local patches in an input image. 
These two parts share the same backbone, while the projection heads are different.
% Since an image contains multiple small local patches and, consequently, multiple local feature representations, we propose a precise location-based feature matching mechanism to improve local dense contrastive learning in digital pathology. 
For the rest of the paper, we use $x$ to represent the input image and $x^q$ and $x^k$ to represent the query and key images, respectively. 
When $x^q$ and $x^k$ are two randomly augmented views of the same image, we optimize the network to pull their feature representations closer. 
When $x^q$ and $x^k$ come from two different images, we optimize the network to push their feature representations apart.

\subsection{Global Contrastive Learning}
% Our method is based on Moco-v2~\cite{he2020momentum_moco_v1,chen2020improved_mocov2} as the general framework, however different contrastive methods could be also applied.
As shown in Fig. \ref{fig:framework} (a), in the left branch (query branch), image $x^{q}$ is passed to a backbone $F(\cdot)$ and a global head $H_G(\cdot)$ to produce a global representation $q$ as shown in Eq.~\ref{equ:q}. 
In the right branch (key branch), the same operation is performed on input image $x^k$ to produce a global representation $k$, as shown in Eq.~\ref{equ:q}.
\begin{align}
    q &= H_G(F(x^q))  \label{equ:q};  \hspace{0.25cm} k = H_G(F(x^k)) 
\end{align}
We then calculate the global contrastive loss $\mathcal{L}^G$ between the two global feature $q$ and $k$ as follows:
\begin{align}
    \mathcal{L}^G &= \mathcal{L}_{con}(q, k)
\end{align}
where the $\mathcal{L}_{con}$ represents the vanilla MoCo-v2 loss~\cite{chen2020improved_mocov2}.

% These features $f^{q}_i$ are passed to a global head $H_G(\cdot)$, in which a global average pooling is conducted, to produce a single global feature $q$. 

% Let's assume, without loss of generalizability, $x^{q}$ is a $96 \times 96$ image and by passing it to the $F$, it produces a $3 \times 3$ feature map $f^{q}_{i,j}, i, 1, \dots n$. 
% Each $f^{q}_{i, j}$ corresponds to a $32 \times 32$ patch in the input image $x^{q}$ at spacial location $i, j$.

\subsection{Local Dense Contrastive Learning}

Apart from the global head, we also have a local dense head similar to the DenseCL~\cite{wang2021dense_densecl}. 
As shown in Fig. \ref{fig:framework} (b), in the left query branch, backbone $F(\cdot)$ and a dense head $H_D(\cdot)$ map the input image $x^q$ to a set of local query features:
\begin{align}
    Q = H_D(F(x^q))&= \{Q_i\}, i = 1, \dots, n
\end{align}
where $n$ is the number of local features. Similarly, the right key branch produces a set of local key features:
\begin{align}
    K = H_D(F(x^k))&= \{K_i\}, i = 1, \dots, n
\end{align}
Each feature $Q_i$ or $K_i$ in the two sets corresponds to a local patch in the original image $x$; we use $P_i^q$ and $P_i^k$ to represent such regions. 
As shown in Fig. \ref{fig:framework} (c), assuming $x^{q}$ is a $96 \times 96$ patch and $H_D(F(x^q))$ outputs a $3 \times 3$ feature map $\{Q_i\}$, $i = 1, \dots, 9$, each $Q_i$ corresponds to a $32 \times 32$ $P_i^q$ patch in $x^q$.

We calculate a local dense contrastive loss $\mathcal{L}^{L}$ between the two groups of local features $Q$ and $K$ as:
\begin{align}
    \mathcal{L}^{L} &= \mathcal{L}^{pre}(Q, K) + \mathcal{L}^{pre}(K, Q)
\end{align}
where the loss is calculated by matching both $Q$ to $K$ and $K$ to $Q$ and $\mathcal{L}^{pre}(\cdot)$ is the precise location-based feature matching loss we will introduce in the next subsection.
% where $\mathcal{L}^{pre}(Q, K)$ represents the matching loss calculated by matching $Q$ to $K$ and $\mathcal{L}^{pre}(K, Q)$ represents the precise matching loss calculated by matching $K$ to $Q$.
% where the loss is calculated by matching both $Q$ to $K$ and $K$ to $Q$.

\subsection{Precise Location-based Feature Matching}

The key problem in the local dense branch is that random zooming and cropping augmentations lead to spatially mismatched features. 
For example, as shown in Fig. \ref{fig:framework} (c), if the augmentation operation does not contain any zooming or cropping, $Q_{i}$ should precisely match to $K_{i}$, since they represent the same $32 \times 32$ patch context ($P_i^q = P_i^k, \forall i$). 

However, zooming and cropping are among the key augmentations in contrastive schemes~\cite{he2020momentum_moco_v1,chen2020simple_simclr}. To solve this problem, in this study we propose a method to address this limitation. 
As demonstrated in Fig. \ref{fig:framework} (d), when augmentation operations include zooming and cropping, $Q_{1}$ and $K_{1}$ are spatially mismatched, since the represented regions are different.
Instead, $Q_{1}$ should match entire $K_5$ and part of $K_6$, $K_8$ and $K_9$ in the example presented in Fig. \ref{fig:framework} (d). 
Observing this, 
we use a weighted sum of $K_{5,6,8,9}$ to match $Q_1$, where the weights are calculated from the extent of the overlapping areas between $Q_1$ and $K_{5,6,8,9}$. To achieve this, we define a $n \times n$ overlapping matrix $M$ between two augmentations, as shown in Fig.~\ref{fig:framework}(b).
The elements of $M$ are defined as:
\begin{align}
    M_{i, j} &= A(P^q_i \cap P^k_j)
\end{align}
where $P^q_i$ represents the $i^{th}$ patch in the query augmentation and $A(x)$ is an area function that calculates the area of $x$. \JZ{ $M_{i, j}$ can be easily calculated using the bounding boxes of $P^q$ and $P^k$ generated during data augmentation.}
$M_{i, j}$ represents the overlapped area in the original image $x$ between $i^{th}$ patch in the left query augmentation and $j^{th}$ patch in the right key augmentation.
$M_{i, j}$ can be easily calculated from the position and size of the patches $P^q_i$ and $P^k_j$.

To match the local features $Q_i$ to features $K$, we need to find out the overlapping area between $Q_i$ and all possible $K$. \JZ{This overlapping area is the $i^{th}$ row of the overlapping matrix $M$, thus the multiplication $M_{i,*} \cdot m_K$ represents the weighted sum of all $K$ overlapped with $Q_i$. Considering all $i$, it is (in matrix format):}
% \begin{align}
%     K' = r\_set(norm(M \cdot m_K)), \hspace{0.25cm} m_K = [K_1, K_2, \dots, K_n]^\top
% \end{align}
\begin{align}
    K' = {M \cdot m_K}/{||M \cdot m_K||}, \hspace{0.25cm} m_K = [K_1, K_2, \dots, K_n]^\top
\end{align}
\JZ{where $||\cdot||$ represents the row-wise $L^2$ norms. For simplicity, $K'$ is also viewed as a set of its rows $\{K'_{i,*} | i = 1, 2, \dots, n\}$. The same process is repeated for matching $K_i$ to all possible $Q$. }
% where $norm(\cdot)$ represents a row-wise $L^2$ normalization function and $r\_set(m) = \{m_{i,*} | i = 1, 2, \dots, n\}$ is a function to extract all rows of matrix $m$ to form a set. The same process is repeated for matching $K_i$ to all possible $Q$. 
% The only difference is that we should use the $i$-th column as the weight. It should be:
% \begin{align}
%     Q' = r\_set(norm(M^\top \cdot m_Q)), \hspace{0.25cm} m_Q = [Q_1, Q_2, \dots, Q_n]^\top
% \end{align}
\begin{align}
    Q' = {{M^\top \cdot m_Q}/{||M^\top \cdot m_Q||}}, \hspace{0.25cm} m_Q = [Q_1, Q_2, \dots, Q_n]^\top
\end{align}

We then define the weights of each pair matching. 
Matches are not equally important since they have different overlaps.
For example, as shown in Fig. \ref{fig:framework}(d), $Q_1$ is covered by $K_{5,6,8,9}$. 
However, $Q_5$ only overlaps with $K_9$ by a small area and $Q_9$ does not have any overlapping with the $K$. 
We thus define the weight of matching as follows:
\begin{align}
    w_i^q &= \sum_{j} M_{i, j} / A(P^q_i);  \hspace{0.25cm}  w_i^k = \sum_{j} (M^T)_{i, j} / A(P^k_i)
\end{align}
where $A(P^q_i)$ is the area of the $i^{th}$ patch in $x^q$ in the original image $x$, and $A(P^q_i)$ is the area of the $i^{th}$ patch in $x^k$ in the original image $x$, and $w_i^q, w_i^k \in [0, 1]$. The final local loss between $Q$ and $K$ can then be formalized as:
\begin{align}
    \mathcal{L}^{L} &= \mathcal{L}^{pre}(Q, K) + \mathcal{L}^{pre}(K, Q) \\
    &= \frac{1}{\sum_i (w^q_i + w^k_i)} 
    \sum_i( w^q_i \cdot \mathcal{L}_{con} (Q_i, K'_i)
        + w^k_i \cdot   \mathcal{L}_{con} (K_i, Q'_i))
\end{align}
where the $\mathcal{L}_{con}$ represents the contrastive loss function which can be any contrastive loss applicable to the problem.

\subsection{Optimization}

The joint loss $\mathcal{L}$ is defined as the sum of global and local losses as follows:
\begin{align}
    \mathcal{L} &= (1 - \lambda)\mathcal{L}^{G} + \lambda \mathcal{L}^{L} \label{equ:lambda}
    % &= \mathcal{L}_{con}(q, k) + 
    % \frac{1}{\sum_i (w^q_i + w^k_i)} 
    % ( w^q_i * \mathcal{L}_{con} (Q_i, K'_i)
    % + w^k_i * \mathcal{L}_{con} (K_i, Q'_i))
\end{align}
where $\lambda \in [0, 1]$ is a weight hyper-parameter. The parameters in the left query branch $\theta_q$ are optimized end-to-end using the gradients calculated by loss $\mathcal{L}$. 
The parameters in the right key branch $\theta_k$ are optimized using exponential moving average (EMA) as follows:
\begin{align}
    \theta_k &= m \theta_k + (1 - m) \theta_q
\end{align}
where $m \in [0, 1)$ is a momentum coefficient. 
We use the MoCo-v2 contrastive learning loss, InfoNCE~\cite{lai2019contrastive_infonce}, for $\mathcal{L}_{con}$ in our experiments, given by:
\begin{align}
    \mathcal{L}_{con}^{q,k^+, k^-} &= -\log \frac{\exp(q\cdot k^+/\tau)}{\exp(q\cdot k^+/\tau) + \sum_{k^-}\exp(q\cdot k^-/\tau)}
\end{align}
where $q$ is a query representation, $k^+$ represents the positive (similar) key samples, and $k^-$ represents the negative (dissimilar) key samples. $\tau$ is a temperature hyper-parameter. 
A query and a key form a positive pair if they are augmented from the same image, and otherwise form a negative pair.
Our method does not have any requirement on the choice of contrastive loss for $\mathcal{L}_{con}$, making our framework generalizable to other contrastive learning frameworks.

\section{Experiments and Discussion}

\subsection{Datasets}
In our experiments, we use 5 datasets. Two of them, NCT-CRC-HE-100K~\cite{nct} and TCGA-BRCA-100K~\cite{brca}, are used as pretraining datasets. The other three, GlaS~\cite{sirinukunwattana2017gland}, CRAG~\cite{graham2019mild}, and BCSS~\cite{bcss} are used to evaluate downstream tasks.

\paragraphdata{NCT-CRC-HE-100K.} The NCT-CRC-HE-100K dataset~\cite{nct} has 100,000 patches of the size $224\times224$ cropped from 86 H\&E stained colorectal adenocarcinoma cancer and normal tissue slides. All the patches are extracted at 20$\times$ magnification. The patches are annotated into nine classes. However, these patch-level labels are not utilized as we use this dataset for self-supervised pre-training. This dataset is used for pretraining followed by the downstream segmentation on the GlaS and CRAG dataset.

\paragraphdata{TCGA-BRCA-100K.} The TCGA-BRCA dataset~\cite{brca} has 1133 slides from patients diagnosed with either Invasive Ductal
(IDC) or Invasive Lobular Breast Carcinoma (ILC). 
We create a dataset by randomly sampling 100,000 tissue patches at 20$\times$ magnification and denote this dataset as TCGA-BRCA-100K. 
Since this dataset is used for pre-training followed by the downstream segmentation on the BCSS~\cite{bcss} dataset, we ensure the slides for the pre-training and downstream tasks do not have any patient-level overlap.

\paragraphdata{GlaS.} The Gland Segmentation in Colon Histology Images (GLaS) dataset~\cite{sirinukunwattana2017gland} has 165 images of size $775 \times 522$ cropped from 16 H\&E histological sections of stage T3 or T4 colorectal adenocarcinoma. 
The digitization of slides is done at 20$\times$ magnification. 
Each image contains object-instance-level annotations of both the benign and malignant glands.

\paragraphdata{CRAG.} The Colorectal adenocarcinoma gland (CRAG) dataset~\cite{graham2019mild} has 213 images of the size mostly around $1512 \times 1516$ collected from 38 H\&E whole slide images (WSIs). 
The images are sampled at 20$\times$ magnification. 
The annotations include the instance-level segmentation masks of the adenocarcinoma and benign glands in colon cancer.

% \paragraphdata{CRAG} The Colorectal adenocarcinoma gland (CRAG) dataset~\cite{graham2019mild} has 213 images of size $1512 \times 1516$ collected from 38 H\&E WSIs. The images are sampled at 20$\times$ magnification. 
% The annotations include the instance-level segmentation of adenocarcinoma and benign glands in colon cancer.
 
\paragraphdata{BCSS.} The Breast Cancer Semantic Segmentation (BCSS) dataset~\cite{bcss} has over 20,000 semantic segmentation annotations of tissue regions sampled from 151 H$\times$E stained breast cancer images at 40$\times$ magnification from TCGA-BRCA~\cite{brca}. 
The annotations include the segmentation masks of 21 classes, such as Tumor, Stroma, Inflammatory, Necrosis, etc.

\subsection{Implementation Details}
In all the experiments, we use ResNet18~\cite{he2016deep} as our backbone network and generate $n = 7 \times 7 = 49$ local features for $Q$ and $K$. 
We compare pre-training ResNet18 with the multiple baseline methods and our precise location-based SSL method for 200 epochs with a batch size 256.
We use the SGD optimizer with a learning rate of 0.03, weight decay of 0.0001, momentum of 0.9 and apply a cosine annealing learning rate decay policy.
For downstream instance segmentation tasks on the GlaS and CRAG datasets, we use MaskRCNN~\cite{he2017mask} with Resnet18~\cite{he2016deep} backbone. 
We train the network on the CRAG dataset for 15000 iterations and GlaS for 5000 iterations. 
We use a batch size of 16 and a base learning rate of 0.02. 
The other hyperparameters are the default ones in Detectron2~\cite{wu2019detectron2}.
For downstream semantic segmentation on the BCSS dataset, we train a ResNet18 based UNet~\cite{ronneberger2015u_unet} using the AdamW~\cite{loshchilov2018decoupled_adamw} optimizer with a batch size of 32, a learning rate of $5e^{-4}$, and a cosine annealing learning rate decay. 
We use the PyTorch library~\cite{paszke2019pytorch}, adopting the OpenSelfSup~\cite{mmselfsup2021} code base.
We train our models on NVIDIA Tesla A100 and Nvidia Quadro RTX 8000 GPUs.

\begin{table}[t]
\caption{Quantitative results of object detection, instance segmentation, and semantic segmentation. For GlaS and CRAG, model is pre-trained for 200 epochs on  the NCT dataset. For BCSS, model is pre-trained for 200 epochs on randomly sampled patches from TCGA-BRCA. VICRegL$^m$ corresponds to a dense matching extension of VICRegL~\cite{bardes2022vicregl} in MoCo-v2 framework.}
\label{table:result:accuracy}
\begin{center}
\setlength{\tabcolsep}{1.6mm}{

\begin{tabular}{c r r r r r r}
\toprule
\multicolumn{1}{c}{Dataset} & \multicolumn{2}{c}{GlaS} & \multicolumn{2}{c}{CRAG} &
\multicolumn{2}{c}{BCSS}\\
% \midrule
Metric & \multicolumn{1}{c}{$AP_{det}$} & \multicolumn{1}{c}{$AP_{seg}$} & \multicolumn{1}{c}{$AP_{det}$} & \multicolumn{1}{c}{$AP_{seg}$} & \multicolumn{1}{c}{Jaccard} & \multicolumn{1}{c}{Dice} \\
\midrule
% ImageNet-pretrained
%     & 52.3 & 55.3 
%         & 50.0 & 50.3
%             & 0.6529 & 0.7771 \\ 
MoCo-v2
    & 52.3 & 55.3 
        & 50.0 & 50.3
            & 0.6529 & 0.7771 \\ 
w/ $M_{ft}$ (DenseCL~\cite{wang2021dense_densecl})
    & 53.9 & 56.5
        & 52.3 & 52.2 
            & 0.6547 & 0.7778  \\ 
w/ $M_{ft}$ \& $M_{loc}$ (VICRegL$^m$~\cite{bardes2022vicregl})
    & 51.3 & 56.0 
        & 53.5 & 51.1
            & 0.6554 & 0.7783 \\ 
\midrule
Ours ($M_{pl}$)
    &\textbf{ 55.0} & \textbf{57.5}
        & \textbf{54.5} & \textbf{54.0}
            & \textbf{0.6559} & \textbf{0.7787}\\ 
% Ours-att 
%     & \textbf{55.11} & \textbf{57.90}
%         & 53.73 & 52.97
%             & \textbf{0.6559} & \textbf{0.7792} \\
\bottomrule
\end{tabular}
}
\end{center}
\end{table}

\subsection{Results}

We evaluate the performance of our method on the three downstream datasets involving colorectal and breast cancers.
% For the downstream instance segmentation tasks on the GlaS and CRAG datasets, we pre-train the model on the NCT-CRC-HE-100K colon cancer dataset. 
% Whereas for BCSS, we pre-train the model on the TCGA-BRCA-100K breast cancer slides. 
After pre-training, the model is used as the backbone for the downstream segmentation tasks. 
Since the tasks in the GlaS and CRAG datasets involve instance segmentation, we use the COCO-style \cite{lin2014microsoft_coco} metrics to evaluate the model: mean average precision for detection and segmentation, denoted by $AP_{det}$ and $AP_{seg}$, respectively. For the BCSS dataset, we use the Jaccard index, and the Dice score to evaluate the quality of predictions.

\JZ{\noindent\textbf{Segmentation performance evaluation.} }
% \subsubsection{Evaluation of overall performance.} 
\textit{Pre-training:} We use vanilla MoCo-v2 as the base SSL framework. 
To evaluate the dense contrastive learning baselines, we use the feature-based matching DenseCL and location-based matching VICRegL in MoCo-v2. 
DenseCL corresponds to MoCo-v2 w/ $M_{ft}$, whereas MoCo-v2 w/ $M_{ft}$ \& $M_{loc}$ corresponds to adoption of VICRegL~\cite{bardes2022vicregl} in the MoCo-v2 framework (denoted as VICRegL$^m$). 
Our precise location-based matching $M_{pl}$ in MoCo-v2 is denoted in the rest of the paper as MoCo-v2 w/ $M_{pl}$. In Table \ref{table:result:accuracy}, we observe that pre-training MoCo-v2 with dense matching techniques such as $M_{ft}$ or $M_{loc}$ results in better performance.
Our proposed dense matching method $M_{pl}$, unlike $M_{loc}$, uses better zooming and cropping augmentations and thus is a more reliable as a complete dense matching strategy. 
In Table \ref{table:result:accuracy}, we empirically verify this claim by showing a consistent improvement across multiple datasets and downstream tasks. Compared to vanilla MoCo-v2, our method achieves a relative improvement in average precision of 5.2\% and  9\% in detection, 4.0\% and 7.3\% in instance segmentation on GlaS and CRAG, respectively. Compared to DenseCL, we see a consistent relative improvement of 2.0\% and 4.2\% in detection, 1.8\% and 3.5\% in instance segmentation; compared to VICRegL$^m$, we see a relative improvement of 7.2\% and 1.9\% in detection, 2.7\% and 5.7\% in instance segmentation on GlaS and CRAG respectively. For semantic segmentation, compared to vanilla MoCo-v2, relative improvement in Jaccard index is up to 0.45\%. 

\noindent\textbf{Evaluation of generalizability.}
% \subsubsection{Evaluation of generalizability.}
% We test the generalizability of utilizing our precise location-based matching with other SSL frameworks.
We demonstrate the generalizability of our precise location-based matching by incorporating $M_{pl}$ into popular SSL frameworks, including vanilla MoCo-v2~\cite{chen2020improved_mocov2}, VICRegL~\cite{bardes2022vicregl}, and ConCL~\cite{yang2022concl}.
% with $M_{pl}$ adopted dense counterparts. 
All experiments are performed on the GlaS dataset in this study. 
In Table \ref{table:result:generalizability}, we observe that our matching method consistently boosts the performance of all the SSL frameworks. 
This shows that our precise location-based matching can be easily adopted by a diverse set of SSL frameworks to boost the representation learning abilities for the dense prediction tasks such as detection and segmentation.

\begin{table}[t]
\caption{Experiments on the generalizability of our proposed method on the GlaS dataset: 1) MoCo-v2 (global contrastive learning), 2) VICRegL (global + local dense contrastive learning), 3) ConCL (global + clustering-based contrastive learning). 
%w/ denotes with and r/ denotes location-based loss replaced by our precise location-based loss. \JZ{make the table horizontal to save spaces}
}
\label{table:result:generalizability}
\begin{center}
\setlength{\tabcolsep}{1.6mm}{
\begin{tabular}{c r r r r r r}
\toprule
% \multicolumn{1}{c}{Dataset} & \multicolumn{2}{c}{GlaS}\\
% \midrule
CL method 
    & \multicolumn{2}{c}{MoCo-v2~\cite{chen2020improved_mocov2}}
    & \multicolumn{2}{c}{VICRegL~\cite{bardes2022vicregl}}
    & \multicolumn{2}{c}{ConCL~\cite{yang2022concl}} \\

Metric 
& \multicolumn{1}{c}{$AP_{det}$} & \multicolumn{1}{c}{$AP_{seg}$} 
& \multicolumn{1}{c}{$AP_{det}$} & \multicolumn{1}{c}{$AP_{seg}$} 
& \multicolumn{1}{c}{$AP_{det}$} & \multicolumn{1}{c}{$AP_{seg}$} \\

\midrule
vanilla method
    & 52.3 & 55.3 
        & 48.3 & 52.3
            & 56.8 & 58.7 
         \\ 
vanilla method w/ $M_{pl}$
    & \textbf{55.0} & \textbf{57.5}
        & \textbf{49.0} & \textbf{53.8}
            & \textbf{57.2} & \textbf{59.1}
        \\ 
\bottomrule
\end{tabular}
}
\end{center}
\end{table}

\begin{figure}[h]
\begin{center}
\includegraphics[width=0.9\linewidth]{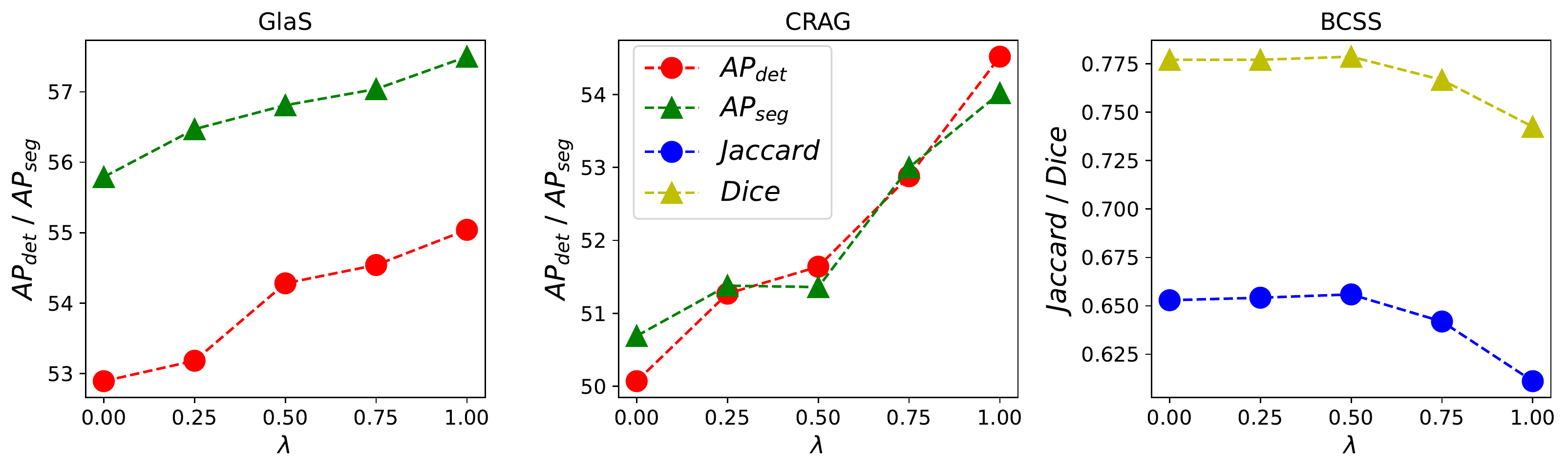}
\end{center}
   \caption{Illustration of the effect of loss weight $\lambda$ on model performance 
   % between our local loss $\mathcal{L}^{L}$ and global loss $\mathcal{L}^{G}$ in equation (\ref{equ:lambda}). 
   The optimal $\lambda$ for the GlaS and the CRAG dataset is $1.0$, and the optimal $\lambda$ for the BCSS dataset is $0.5$.
   }
\label{fig:alpha_ablation}
\end{figure}

\begin{figure}[h]
\begin{center}
%\fbox{\rule{0pt}{2in} \rule{.9\linewidth}{0pt}}
\includegraphics[width=0.95\linewidth]{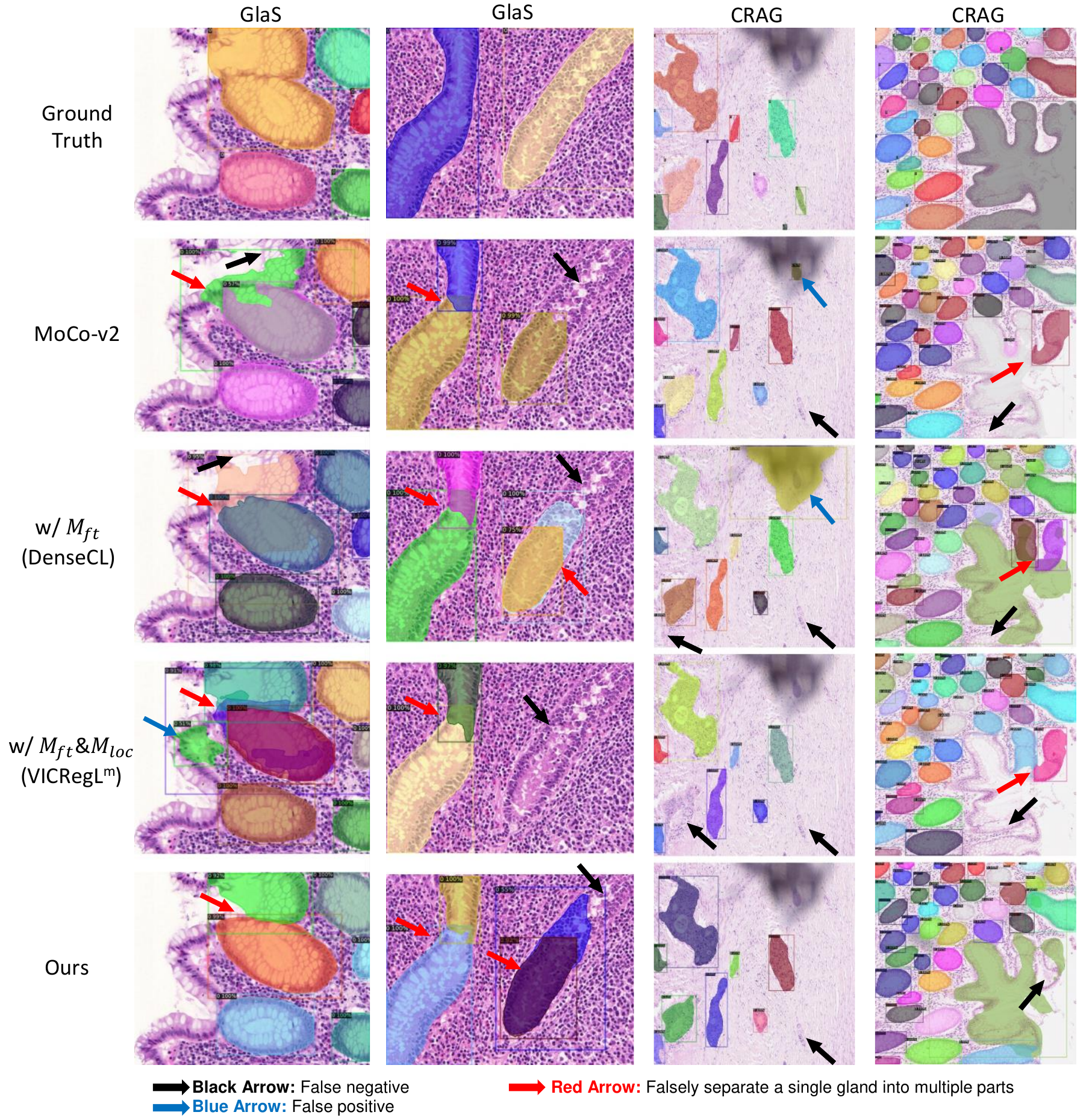}
\end{center}
   \caption{Qualitative comparison on the GlaS and the CRAG datasets. Our method has fewer false negative and false positive segmentations, outperforming other methods.}
\label{fig:vis_seg}
\end{figure}

\noindent\textbf{Ablation study on the hyperparameter $\lambda$.}
% \subsubsection{Ablation study on the hyperparameter $\lambda$.}
To study the optimal weight of the global and the local losses, we perform an ablation study on $\lambda \in \{0.0, 0.25, 0.5, 0.75, 1.0\}$. When $\lambda= 0.0$, the global contrastive loss is used alone for the training, whereas for $\lambda=1.0$, only our proposed local dense loss is used. 
Figure \ref{fig:alpha_ablation} demonstrates that overall our proposed loss boosts the performance of the models on all three datasets. In particular, for instance segmentation tasks (GlaS and CRAG datasets), where we have multiple local objects to segment, our formulation alone provides the best performance and global contrastive schemes may not be so helpful. On the other hand, for semantic segmentation tasks (BCSS dataset), the best performance is achieved when both global and local loss components are combined. Indeed, for such tasks, global interactions of different regions are important to capture different structures. 

\noindent\textbf{Qualitative comparison.} 
% \subsubsection{Qualitative comparison.} 
To qualitatively compare the performance of our method against others, we visualize the segmentation masks and detection boxes in Figure \ref{fig:vis_seg}. Different detection errors (false positives, false negatives and falsely separating a single gland into multiple parts) are indicated by arrows in different colors. 
Overall, our method has fewer false positive and false negative errors, outperforming previous methods and providing more robust segmentations.
%Overall, our method outperforms.... . our method has 
%Compared with MoCo-v2, our method 

% \subsubsection{Visualization of matching.} 
% In order to further show the preciousness of our matching mechanism, we visualize the matched patches using three matching mechanism, feature similarity based matching $M_{ft}$, location-based matching $M_{loc}$ and our precious matching $M_{pl}$.
% As shown in Fig. \ref{fig:matching}, $M_{ft}$ matches visually not similar local features (patches) together. 
% $M_{loc}$ 

% \begin{figure}
% \begin{center}
% %\fbox{\rule{0pt}{2in} \rule{.9\linewidth}{0pt}}
% \includegraphics[width=0.75\linewidth]{imgs/matching_vis.pdf}
% \end{center}
%    \caption{Local feature matching using three methods. Green and orange arrows indicate the matching.
%        (a) Feature similarity based matching ($M_{ft}$), used by DenseCL~\cite{wang2021dense_densecl}. The network pairs visually not similar local features (patches) together.
%        (b) Location-based matching ($M_{loc}$), used by VICRegL~\cite{bardes2022vicregl}. It matches one local feature (patch) to only one feature.
%        (c) Precise location based matching ($M_{pl}$). It preciously matches local features (patches) to multiple features.
%        Blue and red boxes on the left shows augmentation position on the input image. 
%    }
% \label{fig:matching}
% \end{figure}
\section{Conclusion}

In this paper, we introduced a precise location-based matching for SSL frameworks that matches a local patch in a view to multiple corresponding overlapping patches in the other view. 
We applied our proposed matching on two pre-training datasets and evaluated on three downstream tasks.
Our method consistently outperforms state-of-the-art local matching strategies, showing substantial improvement in average precision in both detection and instance segmentation.
Moreover, by using our matching mechanism, the average precision in detection and segmentation was improved in the three popular contrastive learning frameworks, demonstrating the method's generalizability.
Our proposed approach shows the promising potential of local matching in self supervised learning. In future work, we will perform extensive cross-validation on the current datasets and further explore better matching mechanisms and their application to a diverse set of computational pathology tasks.

% \subsubsection{Acknowledgements} This work was supported by xxx.
%the ANR Hagnodice ANR-21-CE45-0007, the NSF IIS-2123920 award, Stony Brook Cancer Center donors Bob Beals and Betsy Barton  as well as the Partner University Fund 4D Vision award.

\subsubsection{Acknowledgements} This work was partially supported by the ANR Hagnodice ANR-21-CE45-0007, the NSF IIS-2212046, the NSF IIS-2123920, the NCI UH3CA225021, the NIH 1R21CA258493-01A1, the Stony Brook Provost Venture Fund (ProFund) and generous donor support from Bob Beals and Betsy Barton. 
%ANR Hagnodice ANR-21-CE45-0007, the NSF IIS-2123920 award, Stony Brook Cancer Center donors Bob Beals and Betsy Barton  as well as the Partner University Fund 4D Vision award.

\bibliographystyle{splncs04}
\bibliography{bibliography.bib}
\end{document}